\documentclass[journal]{IEEEtran}

\usepackage[pdftex, pagebackref, colorlinks]{hyperref}
\usepackage{bookmark}

\usepackage{graphicx}
\usepackage{overpic}
\usepackage{tabularx}

\usepackage{amsmath,amssymb} 
\usepackage{algorithm,algpseudocode}
\usepackage{bm}
\usepackage{multirow}

\begin{document}

\title{LCNN: Low-level Feature Embedded CNN for Salient Object Detection}

\author{Hongyang~Li,~\IEEEmembership{Student Member,~IEEE,}
        Huchuan~Lu,~\IEEEmembership{Senior Member,~IEEE,}
        Zhe~Lin,~\IEEEmembership{Member,~IEEE,}
        Xiaohui~Shen,~\IEEEmembership{Member,~IEEE,}
        and~Brian~Price,~\IEEEmembership{Member,~IEEE}
}

\markboth{
}%
{Li \MakeLowercase{\textit{et al.}}: Saliency Detection via Label Propagation}

\maketitle

\begin{abstract}
In this paper, we propose a novel deep neural network framework embedded with low-level features (LCNN) for salient object detection in complex images.
We utilise the advantage of  convolutional neural networks to automatically learn the high-level features that capture the structured information and semantic context in the image.
In order to better adapt a CNN model into the saliency task, we redesign the network architecture
based on the small-scale datasets.
Several low-level features are
extracted, which can effectively capture contrast and spatial information
in the salient regions, and incorporated to compensate with the learned high-level
features at the output of the last fully connected  layer.
The concatenated feature vector is further fed into a  hinge-loss SVM detector in a joint discriminative learning manner and
the final saliency score of each region within the bounding box is obtained by the linear combination of the detector's weights.
Experiments on three challenging benchmarks (MSRA-5000, PASCAL-S, ECCSD) demonstrate our algorithm to be effective and superior
than most low-level oriented state-of-the-arts in terms of P-R curves, F-measure and mean absolute errors.

\end{abstract}

\begin{IEEEkeywords}
Convolutional Neural Networks, Feature Learning, Saliency Detection.
\end{IEEEkeywords}

\IEEEpeerreviewmaketitle

\setlength{\parskip}{0.2em}

\section{Introduction}

\IEEEPARstart{H}{umans} have the capability to quickly prioritize external visual stimuli and
localize
interesting regions in a scene.
In recent years, visual attention has become an important research problem in both
neuroscience and computer vision.
One  focuses on eye fixation prediction  to
investigate the mechanism of human visual systems \cite{Judd_2009} whereas the other
concentrates on salient object detection  to accurately identify a region of interest \cite{FT}.
Saliency detection has served as a pre-processing procedure for many vision tasks, such as collages \cite{CA}, image compression \cite{compression}, stylized rendering \cite{RC}, object recognition \cite{recognition}, 
image retargeting \cite{HPS}, etc.

\begin{figure}[t]
\begin{center}
\includegraphics[width=8.8cm]{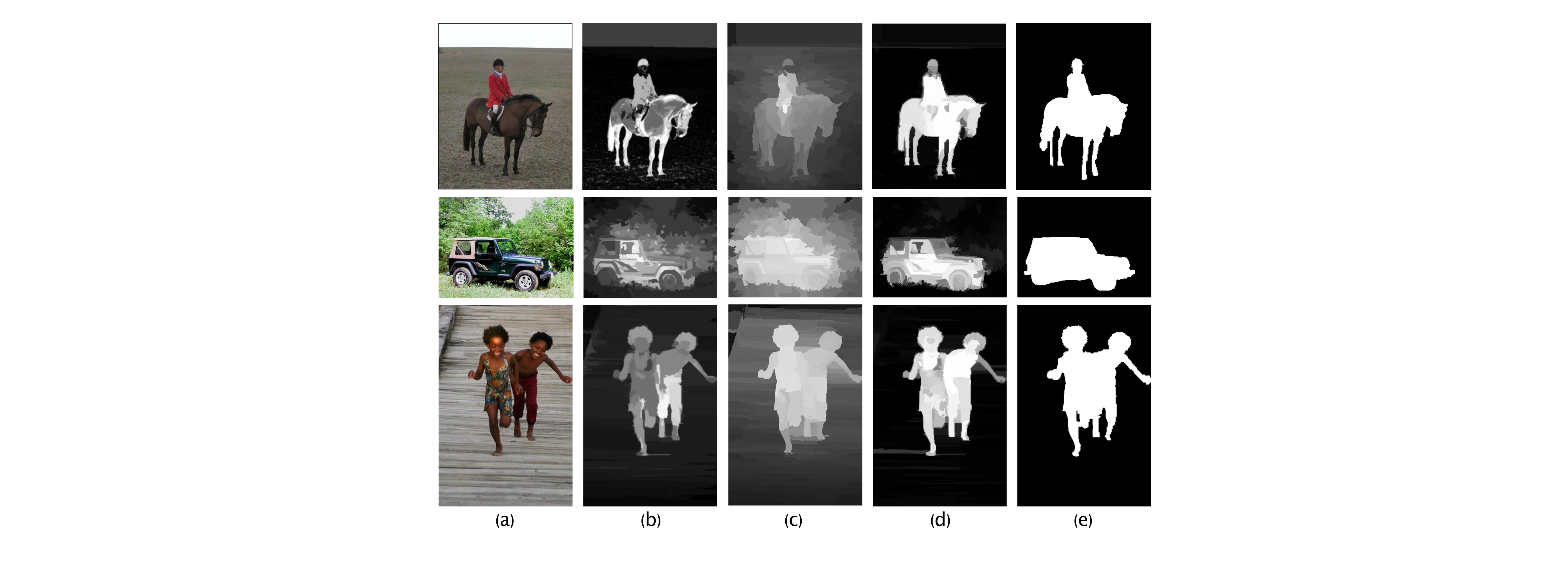}
\end{center}
\caption{Saliency detection results by different methods.
(a) input images; (b) low-level contrast features by \cite{RC};
(c) low-level priors with high-level objectness cues by \cite{SVO};
(d) our LCNN algorithm, which combines high-level features
embedded with low-level priors learned by CNN; (e) ground truth.
}\label{forehead}
\end{figure}

In this work, we focus on accurate saliency detection.
Recently, many low-level features directly extracted from images have been explored.
It has been verified that colour contrast is a primary cue for
obtaining satisfactory
results \cite{SF,RC}.
Other representations based on the low-level features try to exploit
the intrinsic textural difference between the foreground and
background, including focusness \cite{UFO}, textual distinctiveness \cite{TD13}, and structure descriptor \cite{PISA}.
They perform well on simple benchmarks, but can still struggle in images of complex scenarios
since semantic context hidden in the image cannot be effectively captured by hand-crafted low-level priors (see Figure \ref{forehead}(b)).

Due to the shortcomings of low-level features,
several methods have been proposed recently
to incorporate high level features \cite{LR,UFO}.
One type of such representations that can be employed is the notion of objectness \cite{objectness},
\textit{i.e.}, how likely a given region is an object.
For instance, Jiang \textit{et al.} \cite{UFO} computes the saliency map by combining objectness values of 
the candidate windows.
However, using the existent foreground detectors \cite{BING,gop} directly to compute saliency may produce unsatisfying results in complex scenes when  the objectness
score fails to predict  true salient object regions (see Figure \ref{forehead}(c)).

The classic convolutional neural network paradigm \cite{AlexNet,rcnn} has demonstrated superior performance  in image classification and  detection on the challenging databases with complex background and layout in the images (for instance, PASCAL and  ImageNet),
which arises from its ability
to automatically learn high-level features via a layer-to-layer propagation.
This is fundamentally different from previous `objectness' work combining
low-level priors.
Due to the different application background and the scale of datasets, however, a successful adaption of deep model to saliency detection requires a smaller architecture design, a proper definition of the training examples, some refinement scheme such as a low-level feature embedded network, etc.

\begin{figure*}[t]
\begin{center}
\includegraphics[width=17.5cm]{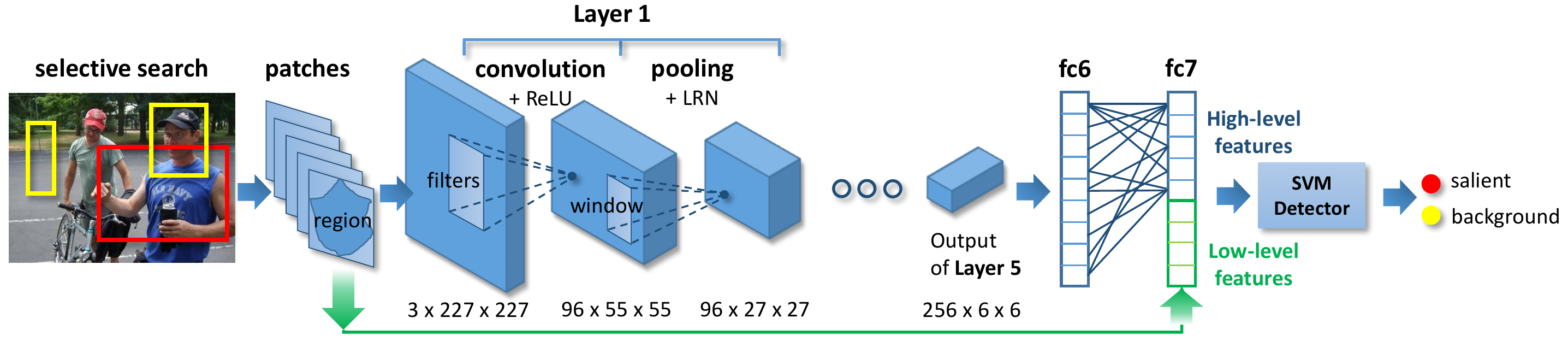}
\end{center}
\caption{Pipeline of the low-level feature embedded deep architecture (LCNN).}\label{pipeline}
\end{figure*}

In this paper, we formulate a novel deep neural network with low-level feature embedded, namely \textbf{LCNN}, which simultaneously leverages the advantage of CNN to capture
the high-level features and that of the contrast and spatial information in  low-level features. To further facilitate the
discriminative characteristics of the network, we combine those extracted features in a joint learning manner via the hinge-loss SVM detector. Figure \ref{forehead}(d) shows the superior advantage of such a deep architecture design, where traditional low-level oriented  method \cite{RC} or high-level objectness-guided algorithm \cite{SVO} fails to detect the salient regions in the complex image scenarios (for example, the salient region has similar colour or texture appearance with the background or it is surrounded by the complicated background).

Figure \ref{pipeline} depicts the general pipeline of our method.
First, a set of candidate bounding boxes with internal region masks are generated by the selective search method \cite{selective_search}; Next, the warped patches are fed into the
deep network to extract high-level features. We make amendments of the classic CNN architecture 
for adaption to the saliency detection problem; Third, a series of simple and effective low-level descriptors are extracted from the regions within each bounding box;
Finally, the concatenated feature vector 
is fed as input  to
 the discriminative SVM detector and the saliency map is generated from the summation of the detector's confidence score.
The experimental results show that the proposed method
achieves superior performance in various evaluation metrics
against the state-of-the-art approaches on three challenging benchmarks.

The rest of our paper reviews related works in section \ref{hell},
describes in detail our CNN framework in section \ref{alg} and low-level feature embedded scheme in section \ref{lcnn}, verifies the proposed model in section \ref{experiment} and concludes the work in section \ref{conclusion}.
Finally, the results and codes will be shared online upon acceptance.

\section{Related Works}\label{hell}

In this section, we discuss the related saliency detection
methods and their connection to generic object detection
algorithms. In addition, we also briefly review deep neural
networks that are closely related to this work.

Saliency estimation methods can be explored from different
perspectives. Basically, most works employ a bottom-up
approach via low-level features while a few incorporate a
top-down solution driven by specific tasks.
In the seminal work by Itti \textit{et al.} \cite{IT}, center-surround differences across multi-scales of image features are computed to detect local
conspicuity.
Ma and Zhang \cite{03ACMMM/Ma_Contrast-based} utilize color contrast
in a local neighborhood as a measure of saliency.
In \cite{GB},
the saliency values are measured by the equilibrium distribution
of Markov chains over different feature maps.
%
Achanta \textit{et al.} \cite{FT} estimate
visual saliency by computing the colour difference between
each pixel w.r.t its mean.
Histogram-based global contrast and spatial coherence are used in \cite{RC} to detect
saliency.
Liu \textit{et al.} \cite{MSRA_dataset_pami} propose a set of features from both
local and global views, which are integrated by a conditional
random field to generate a saliency map.
In \cite{SF}, two
contrast measures based on the uniqueness and spatial distribution
of regions are defined for saliency detection.
To identify small high contrast regions, \cite{HS} propose
a multi-layer approach to analyse the  saliency cues.
A regression model is proposed in \cite{DRFI} to directly
map regional feature vectors to saliency scores.
Recently,
\cite{RB} present a background measurement scheme
to utilise boundary prior for saliency detection.
Liu \textit{et al.} \cite{PDE} solve saliency detection in a novel
partial differential equation manner, where the saliency of certain seeds
are propagated until the equilibrium in the image is ensured.
In \cite{HCT}, colour contrast in higher dimension space is investigated to diversify
the distinctness among superpixels.

Although significant advances have been made, most of the aforementioned
methods integrate hand-crafted features heuristically to generate the final saliency map,
and do not perform well on challenging benchmarks.
In contrast, we devise a deep
network based method embedded with simple low-level priors (LCNN)
to automatically learn features that disclosure
the internal properties of regions and semantic context in complex scenarios.

\begin{table*}[t]

 \renewcommand{\arraystretch}{1.3}
\caption{Architecture details of the proposed deep networks.
C: Convolutional layer; F: Fully-connected layer; P: Pooling layer; R: Rectified linear unit (ReLU);
N: Local response normalization (LRN);
D: Dropout scheme; Channel: The number of output feature maps;
Padding: The number of pixels to add to each side of the input during convolution.}\label{architecture}

\begin{center}

\begin{tabular}{c||c|c |c|c|c|c|c}
    \hline
Layer  		&  1      			&   2  			& 3 		& 4    		& 5 			& 6 			& 7 			\\ \hline
Type   		& C+R+P+N 			& C+R+P+N 		& C+R 	& C+R+P     & C+R+P     	& F+R+D		& F+R+D 		\\ \hline

Input size  & $227\times227$& $27\times27$&$13\times13$&$13\times13$&$6\times6$&$2\times2$&$512+104$	\\ \hline
Channel		&	$96$				&	$256$	&$384$		& $384$  &	$256$	&	$-$		&	$-$		\\\hline
Filter size &$11 \times 11$	&$5\times5$&	$3\times3$	& $3\times3$	&$3\times3$	&	$-$		&	$-$	\\ \hline
Filter stride &$4$				&	$-$		&	$-$	&	 $-$    &$-$ 		&$-$		&	$-$		\\ \hline
Padding		&$-$					&   $2$ 		&	$1$	&	$1$		&$1$			&$-$		&$-$	\\ \hline
Pooling size &$3 \times 3$		&	$3 \times 3$	&$-$	&$3\times3$	&$3\times3$	&$-$		&$-$	\\ \hline
Pooling stride &$2$				&   $2$    		&$-$&	$2$		&$3$			&$-$		&$-$\\ \hline
\end{tabular}

\end{center}
\end{table*}

Generic object detection methods
aim at generating the locations of all category independent objects in an image and have attracted
growing interest in recent years.
Existing techniques
propose object candidates by either measuring the objectness of an image window \cite{objectness,BING}
or grouping regions in a
bottom-up process \cite{gop}.
The generated object candidates
can significantly reduce the search space of category
specific object detectors, which in turn helps other stages
for recognition and other tasks.
To this end, generic object detection
are closely related to salient object detection.
In \cite{objectness},        saliency score is utilized as objectness measurement to generate
object candidates.
\cite{SVO} use a graphical
model to exploit the relationship of objectness and saliency
cues for salient object detection.
In \cite{SSOS}, a random forest
model is trained to predict the saliency score of an object
candidate.
In this work, we utilise the selective search method \cite{selective_search} to generate a series of potential foreground bounding boxes as a preliminary preparation for the inputs of the deep network.

Deep neural networks have achieved state-of-the-art results in
image classification \cite{Donahue_ICML2014,google}, object detection \cite{szegedy2014going,simultaneous} and scene parsing \cite{scene_parsing_1,recurrent}.
The success stems from the expressibility and capacity of deep architectures
that facilitates learning complex features and models to account for interacted relationships
directly from training examples.
Since DNNs mainly take image patches as inputs,
they tend to fail in capturing long range label dependencies for scene parsing as well as saliency detection.
To address this issue, \cite{recurrent} use a recurrent
convolutional neural network to consider large contexts.
In \cite{scene_parsing_1}, a DNN is applied in a multi-scale manner
to learn hierarchical feature representations for scene labeling.
We propose a revised CNN pipeline with low-level feature embedded to consider the
label (region) dependencies based on contrast and spatial descriptors, which is of vital importance in the saliency detection task.

\section{CNN based Saliency Detection}\label{alg}

\begin{figure*}[t]
\begin{center}
\includegraphics[width=0.9\textwidth]{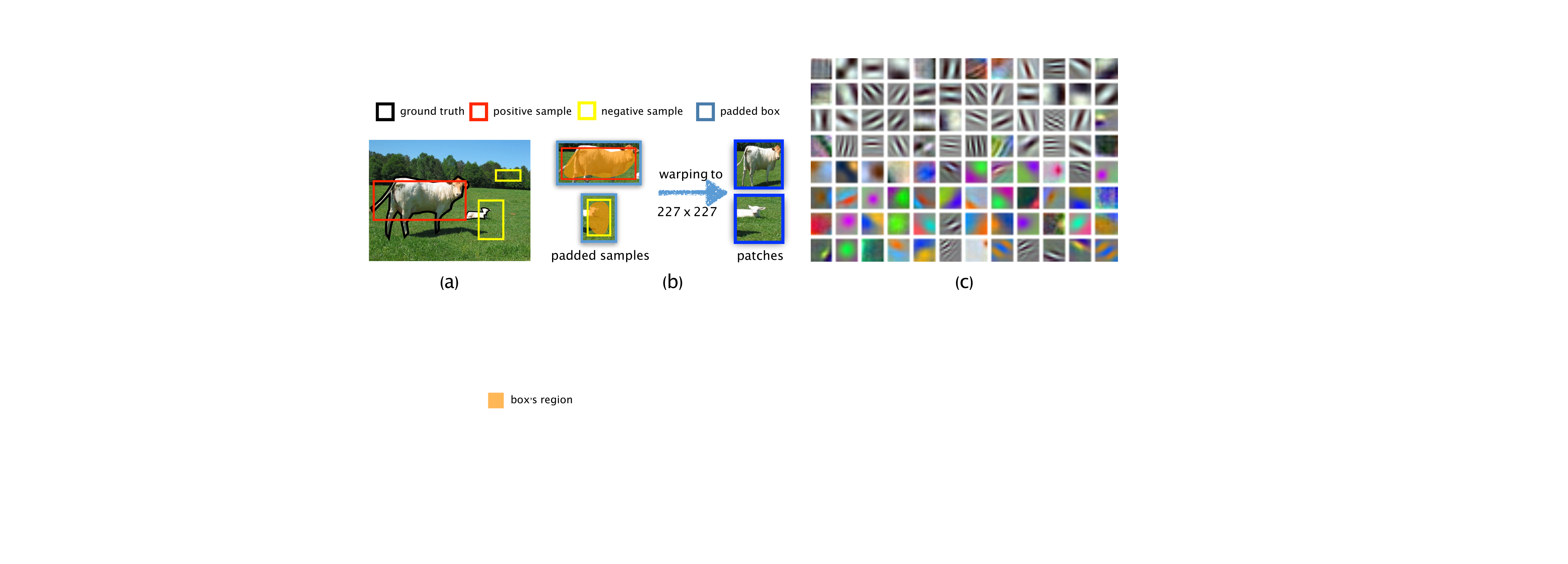}
\end{center}
\caption{(a) Illustration of labelling; (b) Generation of patches. Note that the orange region inside each padded sample is the `cell unit' in our task, which means we use it to extract low-level features and compute saliency;  (c) Visualization of the 96 learned filters in the first layer.
}\label{example}
\end{figure*}

The motivation of applying CNN to saliency detection is that
the network can automatically learn structured and representative features via a layer-to-layer hierarchical propagation
scheme, where we do not have to design complicated hand-crafted features.
The key points to make CNN work for saliency are (a): redesigned network architecture, which means, unlike \cite{rcnn} on the ImageNet \cite{imagenet}, too many layers or parameters will burden the computation in a relatively small-scale saliency dataset; (b): proper definition of positive training examples, that is to say, considering the size of various (maybe multiple) salient object(s), how to define a positive region within the box compared with the ground truth; (c) how to add some 
`refinement' scheme at the output of the last layer to better fit in the accurate saliency detection. Through section
\ref{3_1} to \ref{CNN_saliency}, we will disclosure the solutions of the aforementioned issues respectively.

\subsection{Network architecture}\label{3_1}
The proposed CNN consists of seven layers, with five convolutional layers and two fully
connected layers. Each layer contains learnable parameters
and consists of a linear transformation followed by a nonlinear mapping,
which is implemented by rectified linear units (ReLUs) \cite{AlexNet} to accelerate the training process.
Local response normalization (LRN) is applied to the first two layers
to help generalization.
Max pooling is applied to all convolutional layers except for the third layer to ensure
translational invariance.
The dropout scheme is utilized after the first and the second fully connected layers
to avoid overfitting.
The network takes as input a warped RGB image patch of size $227 \times 227$, and
outputs a 512-dimension feature vector for the SVM detector\footnote{
In the original CNN framework, layer 7 outputs the same feature length (1024-dimension) as layer 6 does.
In order to better balance between high-level and low-level features, we reduce the output number of layer
7 to 512-dimension. Note that in latter experiments without the low-level feature embedded architecture,
layer 7 still outputs a 1024-dimension feature vector.

}. The detailed architecture of the network is shown in Table \ref{architecture}.

To generate the squared patches both for training and test, we first use the selective search method \cite{selective_search} to propose
around 2,000 boxes, each of which also includes the region mask segmented in different color spaces by \cite{pff_segmentation}.
Note that we take a preliminary selection scheme to filter out small boxes or those whose region mask accounts for
little area w.r.t. the whole box. Then we warp all pixels in the tight bounding box around it to the required size.
Prior to warping, we pad the box to include more local context as does \cite{rcnn}.

\subsection{Network training}

\textbf{Training data.} To label the training boxes, we mainly consider the intersection between the bounding box and the ground truth mask. A box $B$ is considered as positive sample if it sufficiently
overlaps with the ground truth region $G$: $| B \cap G | \geq  0.7 \times \max(|B|,|G|)$; similarly,
a box is labeled as negative sample if $| B \cap G | \leq  0.3 \times \max(|B|,|G|)$.
The remaining samples labeled as neither positive nor negative are not
used. Following \cite{AlexNet}, we do not pre-process the training samples, except for subtracting the mean values over the training set from each pixel.
The labelling criteria and the process of patch generation are illustrated in Figure \ref{example}(a)-(b).

\textbf{Cost function.} Given the training box set $\{ B_i \}^N$
and the corresponding label set $\{ y_i \}^N$, we use the softmax loss with weight decay as the cost function:
\begin{equation}
L(\bm{\theta})=-\frac{1}{m}\sum_{i=1}^{m}\sum_{j=0}^{1} \delta(y_i, j)\log P(y_i=j |\bm{\theta} )
+ \lambda \sum_{k=1}^{7} \| \textbf{W}_k   \|
\end{equation}
where $\bm{\theta}$ denotes the learnable parameters set
of CNN  including the weights and bias of all layers; $\delta$ is the indicator
function; $P(y_i=j |\bm{\theta} )$ is the label probability of the \textit{i}-th
training example predicted by CNN; $\lambda$ is the weight decay parameter;
and $\textbf{W}_k$ indicates the weight of the \textit{k}-th layer.
CNN is trained using stochastic gradient descent with a
batch size of $m = 256$, momentum of $0.9$, and weight decay of $0.0005$.
The learning rate is initially set to $0.01$ and
is decreased by a factor of $0.1$ when the cost is stabilized.
Figure \ref{example}(c)
illustrates the learned convolutional filters in the first layer,
which capture color, contrast, edge and pattern information of the local neighborhoods.

\begin{figure*}[t]
\begin{center}
\includegraphics[width=0.9\textwidth]{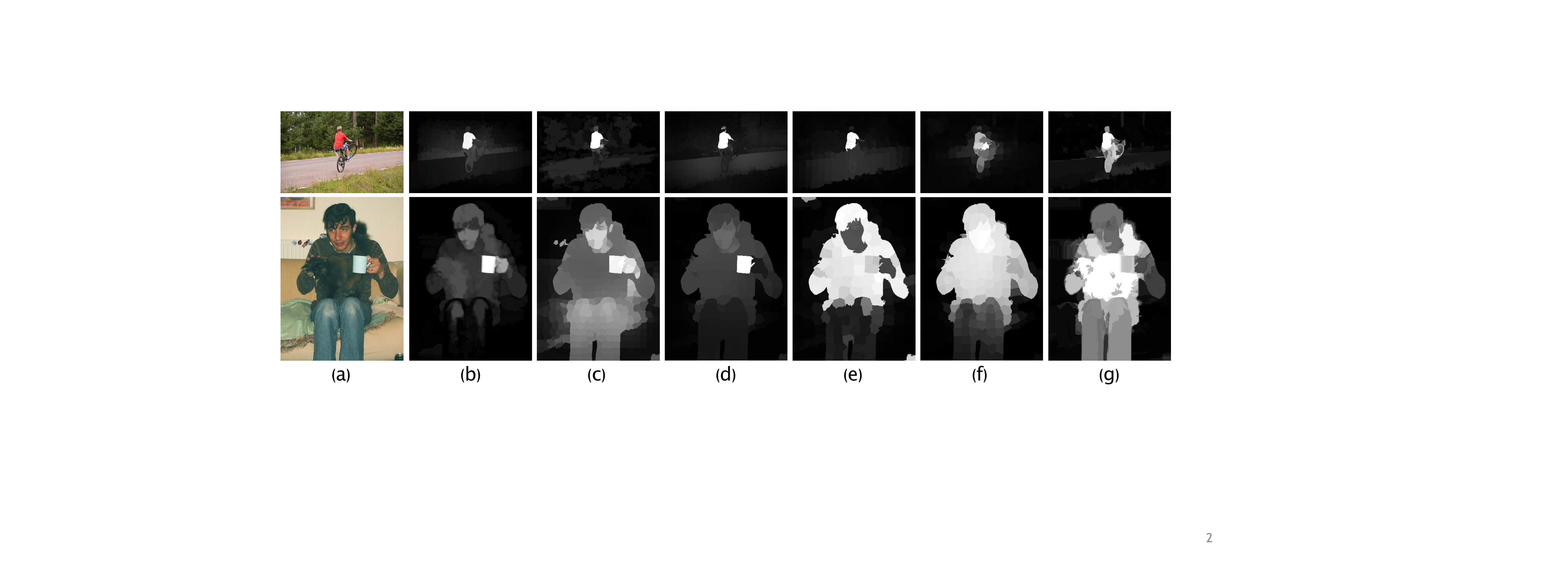}
\end{center}
\caption{Resultant saliency maps of different architecture design. (a) input image;
(b) baseline model; (c) CNN with SVM detector; (d) CNN with spatial descriptors alone;
(e) CNN with contrast descriptors alone; (f) CNN with low-level features (contrast and spatial descriptors together); (g) the proposed LCNN.
}\label{ablation_im}
\end{figure*}

\subsection{CNN for Saliency detection}\label{CNN_saliency}

During the test stage, we feed the trained network with padded and warped patches and predict the saliency
score of each bounding box using the probability $P(y=1 |\bm{\theta} )$. A primitive saliency map is obtained by
summing up the saliency scores of all the candidate regions within the proposed bounding boxes. Figure \ref{ablation_im}(b)
shows the result of directly applying CNN's last layer as the saliency detector to generate saliency maps, which is denoted as the baseline model.
However, as are shown in later experiment (section \ref{ablation}) and  \cite{rcnn}, such a straightforward strategy may suffer from the definition of positive examples used in training the network, which does not emphasise the precise salient localisation within the bounding boxes.

To this end, we introduce a discriminative learning method using the $l_1$ hinge-loss SVM to further classify the extracted high-level features (\textit{i.e.}, the output of layer 7).
The objective function is formulated as:
\begin{equation}
\arg \min_{\textbf{\textit{w}}}\frac{1}{2} \| \textbf{\textit{w}}  \|^2 + C \sum_{i=1}^{N}
\max( 0, 1 - y_i \textbf{\textit{w}}^{T}\textbf{\textit{x}}_{i} )
\end{equation}
where $\textit{\textbf{w}}$ is the weights of the SVM detector and $C$ the penalty coefficient. Here we set
$C=0.001$ to ensure the computation efficiency. The revised saliency score of each bounding box or internal region
is calculated as $ \textbf{\textit{w}}\cdot \textbf{\textit{x}}^{7}+ \textbf{\textit{b}}$, where
$\textbf{\textit{w}}, \textbf{\textit{b}}$ represent the weights and biases of the detector and $\textbf{\textit{x}}^{7}$
being the output feature vector of the fc7 layer.
Figure \ref{ablation_im}(c) depicts the visual enhancement of the saliency maps after enforcing a SVM mechanism, which
can discriminatively choose representative high-level features to determine saliency for the region.

So far, the CNN framework with a SVM detector predicts saliency values based solely on
the automatic learned high-level features,
which can include high-level semantic context in the image via
the box padding and a layer-to-layer propagation scheme. We find by adding some simple low-level priors, such as contrast or geometric information, the CNN framework could obtain much more enhanced results.

%

\section{LCNN: Low-level Feature Embedded CNN}\label{lcnn}

The motivation why high-level feature from CNN alone is not enough can be explained as follows. The CNN-based prediction determines saliency solely based on how a particular sub-region looks like an object bounding box;
the low-level saliency methods  are typically cued on contrast or spatial cues from the global context, which is another valuable information missing in the somewhat `local' CNN prediction.
%
%
%
In this section, we propose a small, and yet effective, set of simple low-level features to compensate with those high-level features in a joint learning spirit. Different from \cite{DRFI} where too many low-level features are proved to be redundant \cite{DRFI_pami}, we use the most common priors, such as colour contrast and spatial properties. To enlarge the feature space diversity, we also explore the texture information in the image by extracting LBP  feature \cite{LBP} and LM filter banks \cite{LM}.

\subsection{Exploring low-level features}

The proposed 104-dimensional low-level features covers a wide diversity from the colour and texture contrast of a region to the spatial properties of a bounding box.
First, given a region $R$ within the bounding box generated by the selective search method and using the RGB colour space as an example, we compute its RGB histogram $\textbf{h}_{R}^{RGB}$, average RGB values $\textbf{a}_{R}^{RGB}$ and RGB color variance $var_{R}^{RGB}$ over all the pixels in the candidate region.
Then, in order to characterize the texture feature of the region,
we calculate the max response histogram of LM filters $\textbf{h}_{R}^{LM}$, the histogram of LBP feature $\textbf{h}_{R}^{LBP}$, the  absolute response of LM filters
$\textbf{r}_{R}$, as well as the variance of the LBP feature $var_{R}^{LBP}$ and the LM filters $var_{R}^{\textbf{r}}$.
Furthermore, we define the border regions of 20 pixels width in four directions of the image
as boundary regions\footnote{
Since the boundary regions in different directions may have different appearance, we compute
their measurements separately. For notation convenience, we denote the feature vectors of the boundary regions
in each direction with a uniform subscript $B$.

}
and compute the measurements  $\textbf{h}_{B}^{CS}$, $\textbf{h}_{B}^{TX}$, $\textbf{a}_{B}^{CS}$, $\textbf{r}_{B}$ in a similarly way as defined above. Also we consider the colour histogram $\textbf{h}_{I}^{CS}$ of the entire image in three colour spaces.
Here $CS$ denotes the three colour spaces and $TX$ represents the two texture features extracted by LBP and LM.

\begin{table*}

 \renewcommand{\arraystretch}{1.5}  
\caption{The detailed description of low-level features.
$\textbf{r}$ denotes the absolute response of LM filters.
%
%
$d(\textbf{a}_1, \textbf{a}_2) = ( | a_{11}-a_{21}|, \cdots, |a_{1k}-a_{2k}| )$, where $k$ is the feature dimension of vector $\textbf{a}_1$ and $\textbf{a}_2$; $\chi^2(\textbf{h}_1, \textbf{h}_2) = \sum_{i=1}^{b}
\frac{2( h_{1i}-h_{2i}  )^2}{h_{1i}+h_{2i}  }$ with $b$ being the number of histogram bins.
}\label{low_feat}

\begin{center}

\begin{tabular}{c|c|c|c  ||  c|c|c|c}
\hline
\multicolumn{4}{c||}{Contrast Descriptors (color and texture)} & \multicolumn{4}{c}{Spatial/Property Descriptors} \\ \hline
Notation     &  Definition & Notation        &  Definition &Notation    &  Definition &Notation  &  Definition \\\hline
$c_1-c_{4}$&$\chi^2(\textbf{h}_R^{RGB},\textbf{h}_B^{RGB})$
& $c_{16}-c_{27}$ &$d(\textbf{a}_{R}^{RGB}, \textbf{a}_{B}^{RGB})$&	
$p_1-p_2$  &  centroid coordinates    	  &
 $p_{22}-p_{24}$        &	$var_R^{RGB}$	\\\hline

$c_{5}-c_{8}$&$\chi^2(\textbf{h}_R^{Lab},\textbf{h}_B^{Lab})$
& $c_{28}-c_{39}$   &$d(\textbf{a}_{R}^{Lab}, \textbf{a}_{B}^{Lab})$ &
  $p_{3}$          &    box aspect ratio      &
  $p_{25}-p_{27}$    &		$var_R^{Lab}$ 		\\\hline

$c_{9}-c_{12}$&$\chi^2(\textbf{h}_R^{HSV},\textbf{h}_B^{HSV})$ &
 $c_{40}-c_{51}  $&       $d(\textbf{a}_{R}^{HSV}, \textbf{a}_{B}^{HSV})$  &
 $p_{4}$  &  box width     &
  $p_{27}-p_{30}$         &	$var_R^{HSV}$ 	\\\hline

$c_{13}$&$\chi^2(\textbf{h}_R^{RGB},\textbf{h}_I^{RGB})$ &
  $c_{52} -c_{55}$ &   $\chi^2(\textbf{h}_R^{LBP},\textbf{h}_B^{LBP})$    &
 $p_{5}$  &  box height     &
         &		\\\hline

$c_{14}$&$\chi^2(\textbf{h}_R^{Lab},\textbf{h}_I^{Lab})$ &
  $c_{56}- c_{59}$  &     $\chi^2(\textbf{h}_R^{LM},\textbf{h}_B^{LM})$  &
 $p_{6}$  &  $var_R^{LBP}$   &
      &	\\\hline

$c_{15}$&$\chi^2(\textbf{h}_R^{HSV},\textbf{h}_I^{HSV})$ &
 $c_{60}-c_{74}  $ &        $d(\textbf{r}_R, \textbf{r}_B)$ &
 $p_{7}-p_{21}$  &  $var_R^{\textbf{r}}$     &
     &		\\\hline

\end{tabular}

\end{center}
\end{table*}

Equipped with the aforementioned definitions and notations, we define a series set of low-level features.
For the contrast descriptors, we introduce the boundary colour contrast by the
chi-square distance $\chi^2(\textbf{h}_R^{RGB},\textbf{h}_B^{RGB})$ between
the RGB histograms of the candidate region and the four boundary regions, and the Euclidean distance
$d(\textbf{a}_{R}^{RGB}, \textbf{a}_{B}^{RGB})$ between their mean RGB values. The rest of the colour or texture contrast between the region and the boundary regions, or the entire image are computed similarly.
For the spatial descriptors, we not only consider the geometric  information of a bounding box, such as the aspect ratio, height/width and centroid coordinates, but also extract the internal colour and texture variance of the candidate region. Note that all the geometric features are normalised w.r.t. the image size.
Finally, all the low-level features are  summarised in Table \ref{low_feat}.

\subsection{LCNN for saliency detection}

We concatenate the low-level feature vector proposed above with the high-level feature vector generated from layer 7 and use them as input of the SVM detector (see Figure \ref{pipeline}). The revised architecture, namely the low-level feature embedded CNN (LCNN), archives better performance than previous schemes.
Note that prior to feeding the concatenated feature into the SVM detector, we pre-process the data by subtracting
the mean and dividing the standard deviation of the feature elements.
The final saliency map follows a similar pipeline as stated in section \ref{CNN_saliency}
and we refine the map on a pixel-wise level using the manifold ranking smoothing \cite{HPS}.

\begin{figure*}[t]

\begin{center}
\includegraphics[width=\textwidth]{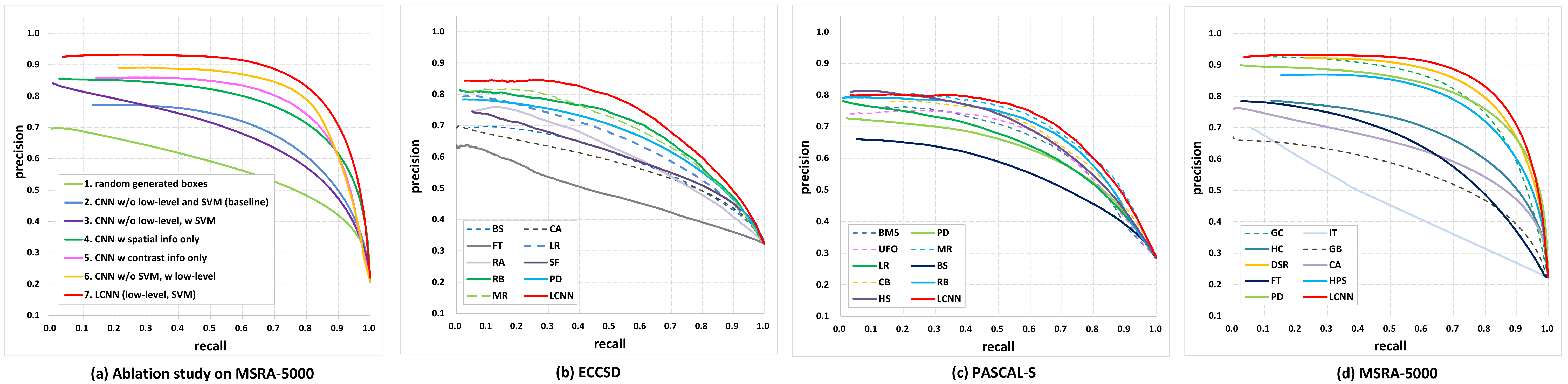}
\end{center}
\caption{Ablation study on MSRA-5000 test dataset and quantitative comparison to previous methods on three benchmarks.}\label{quantitative_fig}

\end{figure*}

Figure \ref{ablation_im}(d)-(f) illustrates the different effects of low-level features. We can see that
the contrast descriptors (row e) play a more important role than the spatial descriptors (row d) as the former considers the appearance distinction between the region and its surroundings.
A combination of the low-level features into the CNN framework (row f)
can effectively facilitate the accuracy of saliency detection since the low-level priors
can catch up the distinctness between the salient regions and the image boundary (usually indicating the background in most cases.).
Furthermore, as Figure \ref{ablation_im}(g)
suggests, our final scheme (LCNN), which includes the SVM detector based on the low-level feature embedded deep network, can take advantage of both low-level priors and discriminative learning detector.
Note that the bicycle and the person's legs are effectively detected in such a framework whereas previous schemes fail to detection them in some way. Figure \ref{quantitative_fig} in section \ref{ablation} proves our architecture design in a quantitative manner.

\section{Experimental Results}\label{experiment}

In this section, we first describe in details the experiment settings on datasets, evaluation metrics and training environment (\ref{setup}); then the ablation studies are conducted to verify each architecture strategy (\ref{ablation}); finally we compare the proposed algorithm with the current state-of-the-arts both in a quantitative and qualitative manner (\ref{fuck}).

\subsection{Setup}\label{setup}

The experiments are conducted on three benchmarks: MSRA-5000 \cite{MSRA_dataset_pami},
ECCSD \cite{HS} and PASCAL-S \cite{SSOS}.
The MSRA-5000 dataset is widely used
for saliency detection and covers a large variety of image
contents. Most of the images include only one salient object with high contrast to the background.
%
The ECCSD dataset consists of  1000 images with complex scenes
from the Internet and is more challenging.
The newly released PASCAL-S dataset descends from the validation set
of the PASCAL VOC 2012 segmentation challenge. This
dataset includes 850 natural images with multiple complex
objects and cluttered backgrounds.
The PASCAL-S dataset
is arguably one of the most challenging saliency datasets without various design biases
(e.g., center bias and color contrast bias).
All the datasets is bundled with pixel-wise ground truth annotations.

We evaluate the performance using precision-recall (PR)
curves, F-measure and mean absolute error (MAE). The
precision and recall of a saliency map are computed by
segmenting the map with a threshold, and comparing the resultant binary map with the ground truth.
The PR curves demonstrate the mean precision and recall of different saliency maps
at various thresholds.
The F-measure is defined as:
\begin{equation}
F_{\beta}=\frac{(1+\beta^2)Precision \times Recall}{\beta^2Precision + Recall}
\end{equation}
where $ Precision $ and $Recall$ are
computed using twice the mean saliency value of saliency
maps as the threshold, and  $\beta^2$ is set to 0.3.
The MAE is the average per-pixel difference between saliency maps $S$ and the
ground truth $GT$:
\begin{equation}
  MAE=\frac{1}{W\times H}\sum_{x=1}^{W}\sum_{y=1}^{H}| S(x,y)-GT(x,y) |.
\end{equation}
where $W, H$ denotes the width and height of the saliency map, respectively.
The metric takes the true negative saliency assignments into account whereas  the precision
and recall only favour the successfully assigned saliency to the salient pixels \cite{GC}.

Since the MSRA-5000 dataset covers various scenarios
and the PASCAL-S dataset contains images with complex
structures, we randomly choose 2500 images from the MSRA-5000 dataset
and 400 images from the PASCAL-S dataset to train the network.
The remaining images are used for tests.
Both horizontal refection and rescaling ($ \pm 5\% $) are applied to all the  training images to
augment the training dataset. The training process is implemented using the Caffe framework \cite{caffe} and initialised
with default parameter setting as suggested in \cite{AlexNet}.
We train the network for roughly 80 epochs through the training set of 1.3 million samples, which takes three weeks
on a NIVIDIA GTX 760 4GB GPU.

\begin{table*}

 \renewcommand{\arraystretch}{1.5}

\caption{Quantitative results using F-measure (higher is better) and MAE (lower is better). The best three results are highlighted
in \textcolor{red}{red}, \textcolor{blue}{blue}  and  \textcolor{green}{green}, respectively.} \label{quantitative_cmp}

\begin{center}
\begin{tabular}{c||c|c|c|c|c|c|c|c|c|c|c|c |c}
\hline
Dataset     &  Metric 					& GC   & HS  & MR   & PD   & SVO & UFO & HPS & RB  & HCT & BMS & DSR & LCNN \\ \hline
\multirow{2}{*}{ECCSD} & F-measure 	& 0.56\footnotemark
& 0.63 &  \textcolor{blue}{0.70} & 0.58 & 0.24 & 0.64  & 0.60 &  \textcolor{green}{0.67} & 0.64  & 0.62 & 0.61 &  \textcolor{red}{0.71}  \\\cline{2-14}
					& MAE		& 0.22 & 0.23 & \textcolor{green}{0.19} & 0.25 & 0.41 & 0.21  & 0.25 & \textcolor{blue}{0.18} & 0.20  & 0.22 & 0.24 &  \textcolor{red}{0.16} \\\hline
\multirow{2}{*}{PASCAL-S}& F-measure 	& 0.49 & 0.54 & \textcolor{green}{0.60} & 0.53 & 0.27 & 0.55 & 0.52 & \textcolor{blue}{0.61} &  0.54  & 0.58 & 0.57 &  \textcolor{red}{0.65}  \\\cline{2-14}
					& MAE		& 0.25 & 0.25 & \textcolor{green}{0.21} & 0.24 & 0.37 & 0.23 & 0.26 & \textcolor{blue}{0.19} &  0.23  & 0.21 & 0.24 &  \textcolor{red}{0.16} \\\hline
\multirow{2}{*}{MSRA-5000}& F-measure 	& 0.70 & 0.77 & \textcolor{red}{0.79} & 0.71 & 0.30 & 0.77 &  0.71 & \textcolor{green}{0.78 }&  0.77 & 0.75 & 0.76 &  \textcolor{blue}{0.79}  \\\cline{2-14}
					& MAE		& 0.15 & 0.16 & \textcolor{green}{0.13} & 0.20 & 0.36 & 0.15 & 0.21 &  \textcolor{red}{0.11} &  0.14 & 0.16 &  0.14 &  \textcolor{blue}{0.12}\\\hline					
\end{tabular}

\end{center}



\end{table*}
\footnotetext{
Note that we round the values to 2 decimal digits.}

\subsection{Ablation studies}\label{ablation}





Figure \ref{quantitative_fig}(a) investigates the performance distinction  of different architecture designs
on MSRA-5000 test dataset in a quantitative manner.
Note that without a preliminary selective search scheme (line 1),
the network suffers from severe insufficient positive samples during training and lacks a proper foreground `guidance'
to predict saliency during test stage. Also the rough score summation of bounding boxes can only generate fuzzy and blurry saliency maps, which is incapable of conducing a precise salient object detection task.
The baseline model (line 2) takes a primitive architecture of Table \ref{architecture} without the final regression
scheme and the introduction of low-level features.
We can see the performance improves slightly after the incorporation of the SVM detector (line 3), particularly in the range of low recall values.
Line 4-6 investigates the different effects of low-level features. We find that the contrast descriptors (line 5)
plays a more important role to facilitate the saliency accuracy that does the spatial descriptors (line 4); and a combination of both contrast and spatial features (line 6) can effectively enhance the result.
Finally, the SVM detector can discriminatively classify the extracted features into the foreground and the background, thus formulating our final version of the low-level feature embedded CNN architecture (line 7).

\subsection{Performance comparison}\label{fuck}

We compare the proposed method (LCNN) with the traditional low-level oriented
algorithms as well as the newly published state-of-the-arts:
IT \cite{IT}, GB \cite{GB}, FT \cite{FT}, CA \cite{CA}, RA \cite{RA}, BS \cite{BS}, LR \cite{LR},
SVO \cite{SVO}, CB \cite{CBS}, SF \cite{SF},  HC \cite{RC}, PD \cite{PD}, MR \cite{MR}, HS \cite{HS},
BMS\cite{BMS}, UFO \cite{UFO}, DSR \cite{DSR}, HPS \cite{HPS}, GC \cite{GC}, RB \cite{RB}, HCT \cite{HCT}.
We use either the implementations or the saliency maps provided by the authors for pair comparison.

\begin{figure*}
\begin{overpic}[width=\textwidth]{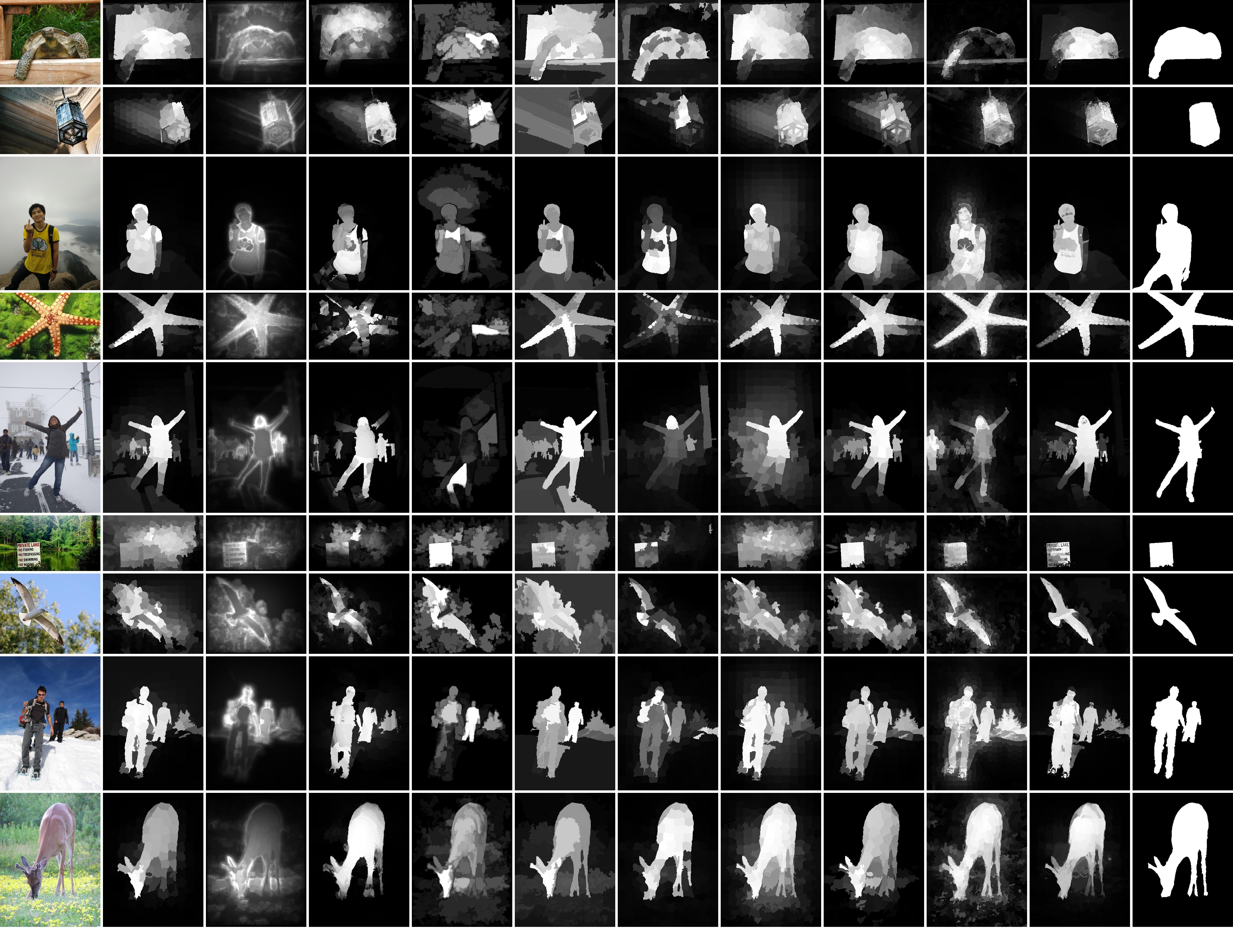}

    \put(1.8, -0.3){
    Input   ~~~~~ MR ~~~~~~ PD ~~~~~~ DS ~~~~~~ BMS ~~~~~ HS ~~~~~~ HPS ~~~~~~ MK
    ~~~~~~ RB ~~~~~ HCT ~~~~ LCNN ~~~~ GT }

\end{overpic}
\caption{Visual comparison of the newest methods published in 2013 and 2014, our algorithm (LCNN) and ground truth (GT).}
\label{visual_compare}
\end{figure*}

Our method performs favourably against the state-of-the-arts on three benchmarks in terms of P-R curves
(Figure \ref{quantitative_fig}), F-measure as well as MAE scores (Table \ref{quantitative_cmp}).
We achieve the highest F-measure value of 0.712, 0.648 and the lowest MAE of 0.161, 0.164 on the ECCSD and PASCAL-S dataset, respectively.
And the performance on the MSRA-5000 dataset is very close to the best method \cite{MR}.
Figure \ref{visual_compare} reports the visual comparison of different saliency maps.
Our algorithm can effectively catch key colour or structure information in complex image scenarios by both learning low-level features and high-level semantic context.

\section{Conclusions}\label{conclusion}

In this paper, we address the salient object detection problem by
learning the high-level features via deep convolutional neural networks and incorporating the low-level features into the deep model to enhance the saliency accuracy. To further catch the discriminant semantic
context in the complex image scenarios, we introduce a hinge-loss SVM detector to better distinguish
the salient region(s) within each bounding box. Experimental results show that our algorithm achieves superior
performance against the state-of-the-arts on three benchmarks.
%
A straightforward extension to our method is to jointly learn global and local saliency context through a novel neural network architecture instead of relying on hand-crafted low-level features, which will be left as our future work.

\bibliographystyle{IEEEtran}
\bibliography{iclr2015}
\end{document}